\pdfoutput=1

\documentclass[11pt]{article}

\usepackage{float}
\usepackage{arydshln}
\usepackage{multirow}
\usepackage[]{acl}
\usepackage{graphicx}
\usepackage{times}
\usepackage{latexsym}

\usepackage[T1]{fontenc}

\usepackage[utf8]{inputenc}

\usepackage{microtype}

\title{Lost in Translation: When GPT-4V(ision) Can't See Eye to Eye with Text \\ A Vision-Language-Consistency Analysis of VLLMs and Beyond}

\author{Xiang Zhang\thanks{$^{*}$Equal contribution.} \\
  University of British Columbia \\
  \texttt{xzhang23@ualberta.ca} \\\And
  Senyu Li\footnotemark[1] \\
  University of Alberta \\
  \texttt{senyu@ualberta.ca} \\\AND
  Zijun Wu \\
  University of Alberta \\
  \texttt{zijun4@ualberta.ca} \\\And
  Ning Shi \\
  University of Alberta \\
  \texttt{ning.shi@ualberta.ca}}

\begin{document}

\maketitle

\begin{abstract}
Recent advancements in multimodal techniques open exciting possibilities for models excelling in diverse tasks involving text, audio, and image processing. Models like GPT-4V, blending computer vision and language modeling, excel in complex text and image tasks. Numerous prior research endeavors have diligently examined the performance of these Vision Large Language Models (VLLMs) across tasks like object detection, image captioning and others. However, these analyses often focus on evaluating the performance of each modality in isolation, lacking insights into their cross-modal interactions. Specifically, questions concerning whether these vision-language models execute vision and language tasks consistently or independently have remained unanswered. In this study, we draw inspiration from recent investigations into multilingualism and conduct a comprehensive analysis of model's cross-modal interactions. We introduce a systematic framework that quantifies the capability disparities between different modalities in the multi-modal setting and provide a set of datasets designed for these evaluations. Our findings reveal that models like GPT-4V tend to perform consistent modalities when the tasks are relatively simple. However, the trustworthiness of results derived from the vision modality diminishes as the tasks become more challenging. Expanding on our findings, we introduce "Vision Description Prompting," a method that effectively improves performance in challenging vision-related tasks.
\end{abstract}

\section{Introduction}
\begin{figure}[t!]
    \centering
    \includegraphics[width=\columnwidth]{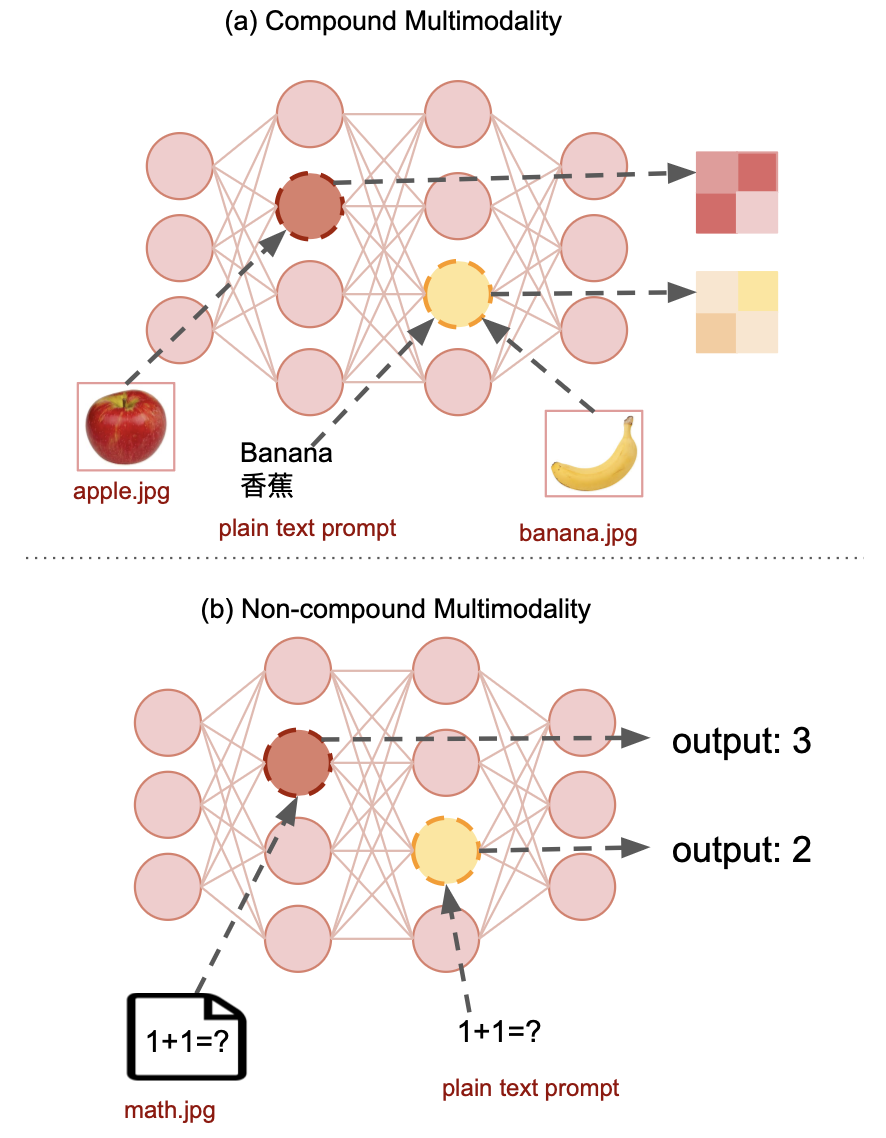}
    \caption{
    A comparison of a proficiently trained compound VLLM versus a poorly trained non-compound VLLM. The compound model integrates representations for the same concept, whereas the non-compound model appears as a rudimentary combination of two distinct models: one for vision and one for language.
    }
    \label{fig:comp}
\end{figure}

Recent large multimodal models have showcased remarkable capabilities in tasks that require the integration of multiple modalities and sources of information~\cite{huang2023chatgpt}. Among these, the performance of Vision Large Language Models (VLLMs)~\cite{zhang2023vision,yang2023dawn} stands out, thanks to the vast amount of image and text data available for training and the rapid progress in both computer vision and language modeling. However, due to the distinct training methodologies employed by these models, such as contrastive learning~\cite{radford2021learning} and embodied image-language modeling~\cite{driess2023palme}, and the varying quality of training data for each modality (e.g., data containing images with varying levels of quality but rich textual content), these networks often exhibit performance disparities across different modalities. This results in either enhanced performance in visual tasks or vice versa, depending on the specific model and training conditions.

Previous research has extensively evaluated the performance of individual modalities in multimodal systems. For instance, ~\newcite{yang2023dawn} conducted a thorough assessment of GPT-4V's vision understanding capabilities, while ~\newcite{chen2023endtoend} analyzed its decision-making abilities. However, assessing a model's performance on each individual modality alone does not fully capture its true multimodal abilities. It is possible for a model to excel in numerous vision tasks but still lag significantly behind in language understanding. This discrepancy can often be attributed to imbalanced training data or differences in the training paradigms for each modality. Moreover, simply testing performance on individual tasks provides no insight into how each modality of the model influences the others. Unfortunately, the cross-modality relationship is frequently overlooked in the aforementioned research.

In this study, our approach transcends the conventional practice of merely evaluating multimodal systems across distinct downstream tasks and subsequently presenting test scores. Instead, our primary focus centers on quantifying the inherent disparities in abilities between different modalities, with a specific emphasis on vision and language, within a VLLM. To enable a comprehensive analysis, we categorize vision-language tasks into two distinct groups: translation variant tasks and translation invariant tasks, contingent upon whether information integrity is maintained during the transition from one modality to another. In each category, we create datasets with task pairs revealing different aspects of vision and language abilities. We use a pairwise consistency metric to measure the disparities between vision and language. Additionally, we investigate the model's internal reasoning for both modalities using tailored prompts, aiming to uncover the network's cross-modality representation and mutual influence. Experiments on GPT4V show that it operates on a highly inconsistent fashion between vision and language, and performance can be largely varied if tasks are done in one modality than the other. Our experiments, conducted on the GPT4V model, reveal a high degree of inconsistency in its performance between vision and language for some given tasks. Moreover, the model's performance exhibits substantial variability depending on whether tasks are conducted primarily in one modality or the other.

The contributions are: (1) We introduce a novel evaluation framework to assess performance consistency across different modalities within a large multimodal system, providing insights into their interactions and differences beyond conventional metrics. (2) We release 6 diverse, meticulously crafted datasets for vision-language consistency testing, fostering future research. (3) Our experiments on GPT-4V reveal a      significant disparity between vision and language abilities within such systems, prompting the introduction of the Vision-Text Prompting (VDP) method as a potential remedy.

\begin{figure}[t]
    \centering
    \includegraphics[width=0.75\columnwidth]{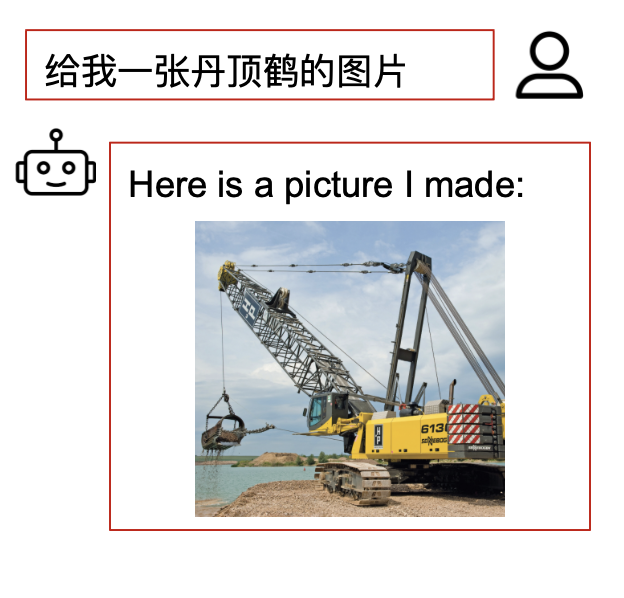}
    \caption{
   An illustration of an error in a specific task using a non-compound network. The prompt, translated from Chinese, requests an image of a "crane" referring unambiguously to the bird. Yet, the provided image depicts the construction vehicle "crane". This suggests the model is relying on translation rather than directly associating words with their intended concepts. The information is lost during the translation and creating ambiguity. 
    }
    \label{fig:mistake}
\end{figure}

\section{A multilingualism view of Vision-Language models types.}
\begin{figure*}[t!]
    \centering
    \includegraphics[width=1\linewidth]{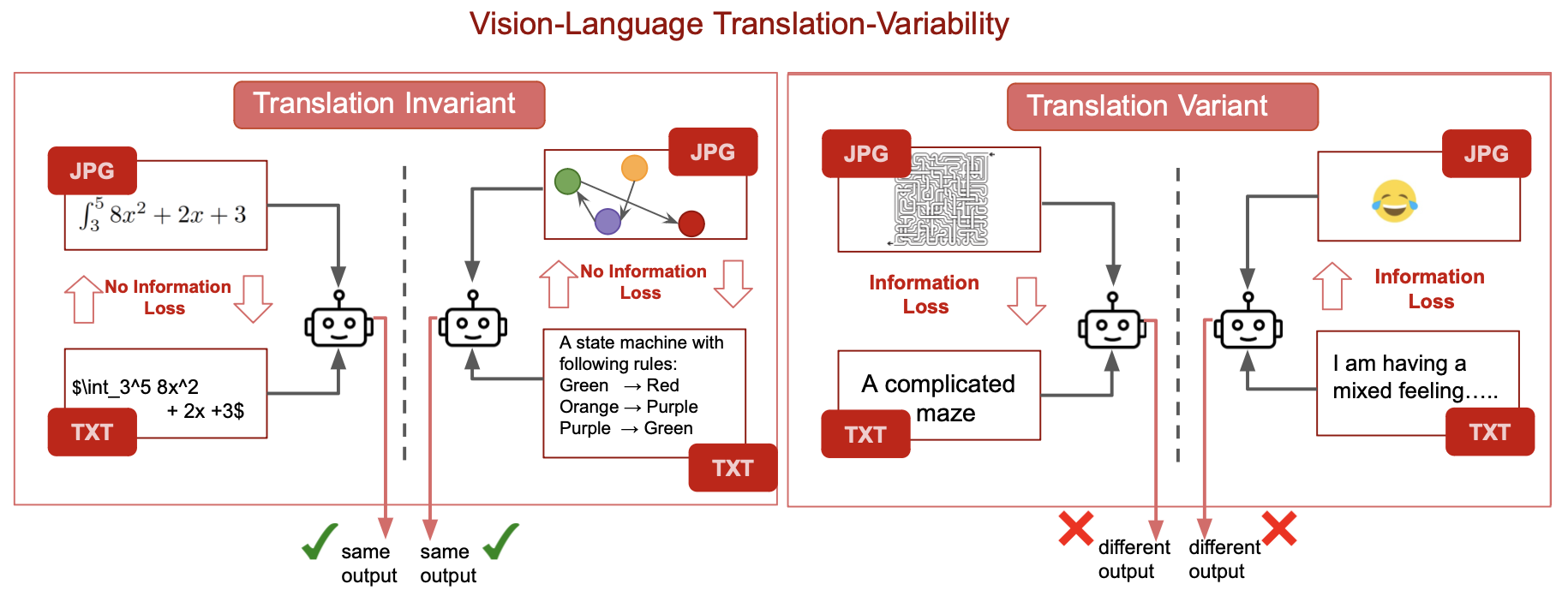}
    \caption{Translation Variant (TV) tasks and Translation Invariant (TI) tasks illustration.}
    \label{fig:TVTI}
\end{figure*}
Differing from the way vision and language information naturally integrate into the human brain during our developmental years, where spoken language often aligns with corresponding visual cues, training models to seamlessly merge these two types of information has proven to be a formidable challenge. A multimodality system that hasn't been trained properly can be thought of as a basic combination of several independent networks, each one handling a different modality. (part b of Figure \ref{fig:comp}). While performance on tasks specific to each modality may be reasonable, it falls far short of achieving an optimal system due to the limited cross-modal information exchange, often necessitating a cumbersome "translation" process. This limitation becomes evident through errors, some of which are illustrated in Figure \ref{fig:mistake}).

To better understand how VLLMs represent multimodal information, we draw insights from recent research on LLMs' multilingual abilities~\cite{zhang2023don}.  We conceptualize Language and Vision as akin to two distinct human languages to be acquired. In linguistic theory, a person is said to possess compound bilingualism~\cite{d1990three,de1991lexical}  if two languages access a shared, learned inner representation~\cite{de1991lexical} and there's no clear distinction between these two languages~\cite{moradi2014investigation,zhang2023bridging}. This phenomenon typically occurs in individuals who grow up in an environment where both languages are acquired simultaneously. Conversely, we refer to a bilingual speaker as having non-compound bilingualism when the two languages exhibit clear distinctions and perform differently for the individual~\cite{jakobovits1968dimensionality}. This latter scenario is often the case for most adult second language learners, as they process the two languages using separate internal systems, and language switching frequently involves an inefficient translation process.

As previous research has indicated~\cite{zhang2023don}, the existence of a compound multilingual Large Language Model is exceptionally rare due to the scarcity of large, paired multilingual text corpora. The training of such models on different language corpora tends to be relatively independent, with minimal to no crossover between languages. As demonstrated by ~\cite{zhang2023don}, these models often primarily handle multilingual tasks by translating into their dominant language. However, it remains uncertain whether the same holds true for VLLMs. This is because the training of multimodal models typically involves distinct techniques~\cite{radford2021learning} and utilizes more paired datasets (e.g. captioned images) specific to the modalities of interest.

Our investigation focuses less on LVLMs' performance in individual tasks and more on assessing their compound-level between vision and language. We hope such an analysis offers valuable suggestions for future research on the usage of these models.
\section{Tasks categorization based on Translation-variability}
To enable cross-modal consistency analysis, we adopt a categorization framework inspired by multilingualism research~\cite{zhang2023don} and introduce a novel paradigm for classifying any given downstream task. We classify tasks as Translation Variant (TV) or Translation Invariant (TI) based on whether the core information essential for task-solving is retained when converting to another modality.

 The concept 'Invariance' pertains to operations that remain consistent in their behavior or response when subjected to input transformations.  In our context, when translating input signals from one modality to another, 'invariance' implies that such translation does not result in any changes to the solver's output.

Formally, considering a transformation function denoted as $f(x)$, which converts an image into a textual description, a vision task is considered 'translation invariant' under the operation of $f(x)$ if, after applying this translation, the multimodal system $g(x)$ is expected to produce the same results as before:
\begin{equation}
    g(x) = g(f(x)) 
\end{equation}
This generalizes to language tasks and tasks of other modalities.

For TI tasks, the task instances typically retain their information when translating between modalities. This preservation ensures that there is no substantial difference in solving the tasks in either modality by a perfect compound system.

Conversely, for TV  tasks, the core information of the task is often lost during translation, creating a hurdle for solving the task in one modality compared to the other. We illustrate examples of TI and TV tasks in Figure \ref{fig:TVTI}, with more comprehensive details to be presented in Section \ref{sec:dataset} concerning dataset construction.

\section{Method}
In this section, our emphasis is on Vision Large Language Models, and we will elaborate on our methodology for quantifying the consistency between the vision and language modalities within a given system.
\subsection{A framework of evaluating cross-modality consistent }

\begin{figure*}[t!]
    \centering
    \includegraphics[width=1\linewidth]{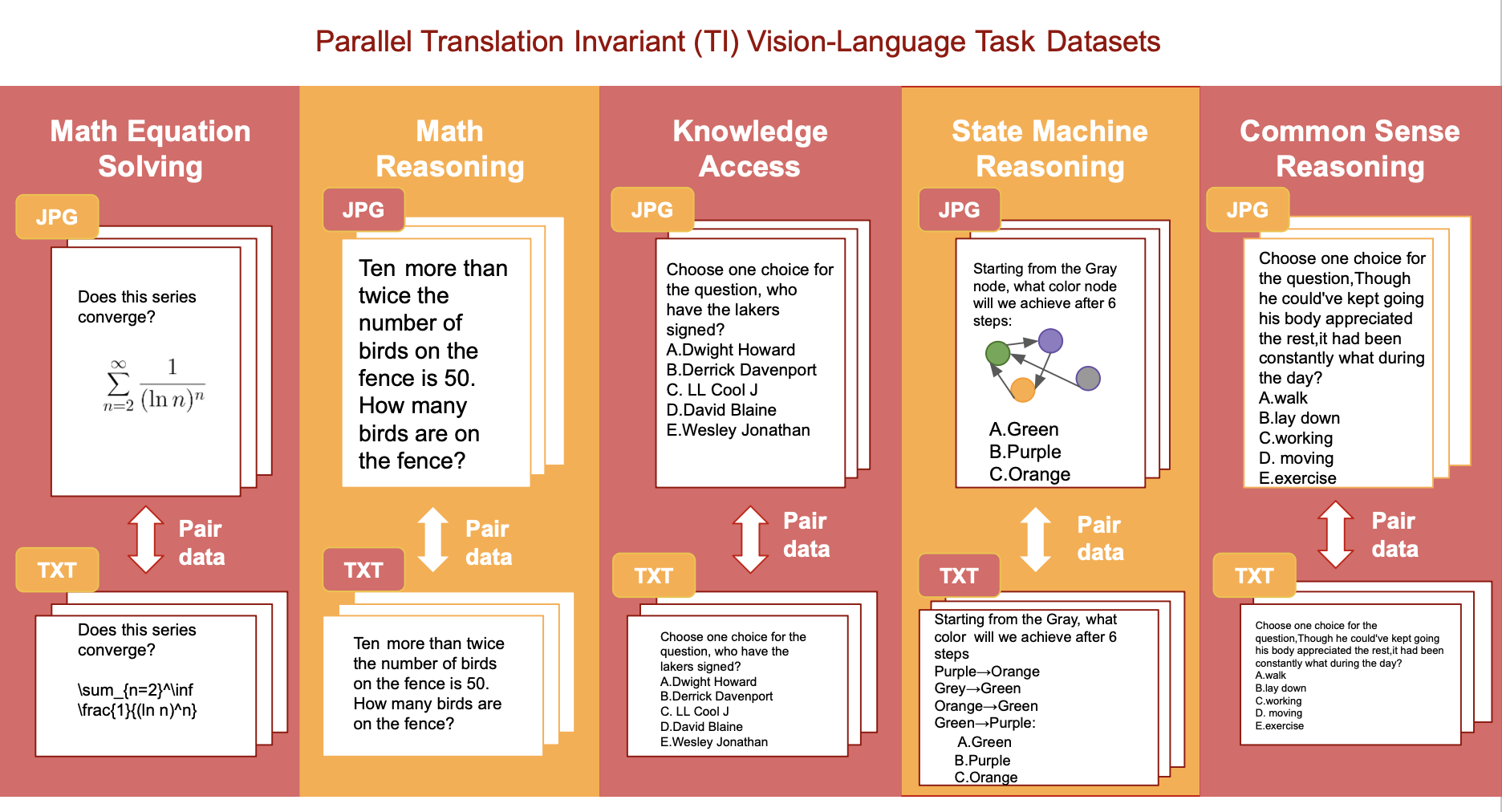}
    \caption{TI Dataset construction. }
    \label{fig:data}
\end{figure*}

As previously mentioned, assessing performance independently on each modality provides limited insights into multimodality models. To address this, we propose an evaluation approach that considers tasks on a pairwise basis. However, quantitatively measuring the disparity in results caused by the model itself can be challenging for TV tasks, as these tasks are expected to yield different outcomes when the same task instance is presented in different modalities. Instead, our focus is on TI tasks, where the disparity in results is not influenced by the modality of information. In TI tasks, we expect the results to be the same if a well-trained compound model is applied. Therefore, by measuring the pairwise results for the same instance, we can assess the disparity between the model's modalities.

For a TI (Translation Invariant) task, our procedure involves obtaining a pair of task instances, as detailed in Section \ref{sec:dataset} and also depicted in Figure \ref{fig:data}, where one instance uses an image and the other employs text for the task description. Importantly, both instances contain the same amount of information necessary for correctly solving the task, as illustrated in the figure~\ref{fig:TVTI} for TI tasks.  We then proceed to prompt the Multimodal system using the image and text separately in two distinct sessions to avoid information crossovers. Subsequently, we calculate the pairwise accuracy by determining the proportion of instances for which both the image results and text results are either correct or incorrect. This metric provides a robust indication of the consistency score across the two modalities.\\

\section{Experiment}

\subsection{An dataset construction approach for Vision-language evaluation under proposed framework }
\label{sec:dataset}
Since there is currently no available parallel vision-language task dataset, we have undertaken the task of constructing our own dataset. We have created  5 Translation Invariant (TI) task datasets and 1 Translation Variant (TV) task dataset.  We have made this dataset publicly accessible \footnote{\url{https://github.com/pooooobzx/Vision-Language-Consistency}}for future research.

\subsubsection{TI tasks dataset.}
\textbf{Math Equation Solving.} In the context of math equation solving, images are particularly effective at representing mathematical equations, offering a clear visualization of complex symbols and notations. To facilitate this task, we have curated a dataset containing images of mathematical questions encompassing various math symbols and expressions. It's important to note that this task is inherently translation-invariant since we can preserve all the necessary information for solving the problems using the \LaTeX \  text format. Consequently, we have paired up all the image-based math questions with their corresponding text representations to create our final dataset for equation solving. An example of this construction can be seen in Figure \ref{fig:data} and data samples can be seen in Appendix \ref{app:MES}. \\
\begin{table*}
    \centering
    \resizebox{0.8\textwidth}{!}{
    \begin{tabular}{|c|cccc|}
    \hline
    \textbf{Translation-Variability} & \textbf{DataSet} & \textbf{Modality} & \textbf{Accuracy} & \textbf{Consitency Score} \\
    \hline\hline
    \multirow{5}{*}{TI} & \multirow{2}{*}{Math Equation solving} & Text & 0.80 & \multirow{2}{*}{0.62} \\
    \cdashline{3-4}
    &  & Image & \ \ \ 0.58 $\color{red}\Downarrow$ & \\
    \cline{2-5}
    & \multirow{2}{*}{Math Reasoning} & Text & 0.96 & \multirow{2}{*}{0.58} \\
    \cdashline{3-4}
    &  & Image & \ \ \ 0.54 $\color{red}\Downarrow$ &  \\
    \cline{2-5}
    & \multirow{2}{*}{Knowledge Access} & Text & 1.00 & \multirow{2}{*}{0.94} \\
    \cdashline{3-4}
    &  & Image & 0.94 &  \\
    \cline{2-5} 
    & \multirow{2}{*}{State Matching Reasoning} & Text & 0.29 & \multirow{2}{*}{0.77} \\
    \cdashline{3-4}
    &  & Image & 0.25 &  \\
    \cline{2-5}
    
    & \multirow{2}{*}{Common Sense Reasoning} & Text & 0.78 & \multirow{2}{*}{0.80} \\
    \cdashline{3-4}
    &  & Image & 0.82 & \\
    \hline\hline
    \multirow{2}{*}{TV} &\multirow{2}{*}{Common Sense VQA} & Text & 0.80 & \multirow{2}{*}{0.87} \\
    \cdashline{3-4}
    &  & Image & \ \ \  0.67 $\color{red}\Downarrow$ &  \\
    
    \hline
    \end{tabular}}
    \caption{Test results for vision-language datasets pertaining to TI and TV tasks. The symbol $\color{red}\Downarrow$ denotes a significant decrease in accuracy (greater than 10\%) when input is in image format. }
    \label{tab:main}
\end{table*}
\textbf{Math reasoning.} For the task of math and logical reasoning, we utilized the GSM8K~\cite{DBLP:journals/corr/abs-2110-14168} dataset, which comprises 8,500 questions (7,500 for training and 1,000 for testing) of this nature. To enable multimodal analysis, we converted each of these questions into a well-formatted image file. The resulting images and the original text files were then paired together to create a parallel dataset, enabling the exploration of this task in both image and text modalities.  An example of this construction can be seen in Figure \ref{fig:data} and data samples can be seen in Appendix \ref{app:MR}.\\
\textbf{Knowledge Access.} To assess the difference in the ability to access factual knowledge between the vision and text modalities, we have developed a Knowledge Access Parallel Dataset using the well-established WebQuestions~\cite{bordes-etal-2014-question} dataset. This dataset comprises 6,642 questions in text format. An example of such a question is "Which country did Justin Biber grows up at?" To enable multimodal analysis, we have once more converted each question into an image file format before pairing it up with the original text, thereby creating a dataset that allows for the exploration of knowledge access in both image and text modalities. An example of this construction can be seen in Figure \ref{fig:data} and data samples can be seen in Appendix \ref{app:knowledge}.\\
\textbf{Common sense reasoning.}To investigate potential differences in common sense reasoning when different modalities are used, we have established a parallel dataset based on the CommonsenseQA~\cite{talmor-etal-2019-commonsenseqa} dataset. This dataset consists of 12,247 multiple-choice questions, each of which has been transformed into an image format, with all answer choices displayed below the questions. We have paired these images with the original text questions, facilitating the examination of common sense reasoning in both text and image modalities. An example of this construction can be seen in Figure \ref{fig:data} and data samples can be seen in Appendix \ref{app:common}. \\
\textbf{State Machine Reasoning.} State machines, which can be effectively visualized as graphs or represented through text with transitional rules, serve as an ideal testbed for assessing the reasoning capabilities of VLLMs. Our approach involves generating images of state machines with varying total numbers of nodes (states). Each node in the state machine is assigned a distinct color and features precisely one outgoing edge, ensuring a unique path and solution.

The questions we formulate are of the form, "Starting from the color grey, after n steps, which color will we end up in?" Here, n is a variable that we select. Additionally, we generate a text version of these state machines by listing out all the transition rules for each arrow. To prevent any form of cheating by looking at the last state in the text, we shuffle the order of the rules. We create state machines with different numbers of states and questions with varying numbers of steps to introduce varying difficulty levels. The text-image pairs corresponding to these state-machine questions constitute our final dataset for state-machine reasoning. An example of this construction can be seen in Figure \ref{fig:data} and data samples can be seen in Appendix \ref{app:state}.

\subsubsection{TV task dataset.}
\textbf{VQA common sense reasoning.} Given that many VQA (Visual Question Answering) datasets involve tasks that cannot be readily converted into meaningful text-based counterparts, such as counting objects in an image or solving mazes, we have created a parallel vision-language VQA dataset using the CommonsenseQA~\cite{talmor-etal-2019-commonsenseqa} dataset.

In this process, for each question in CommonsenseQA, we have replaced the text description of the question with a corresponding image while keeping all other elements unchanged. The original text question from CommonsenseQA is then utilized as the text version of the dataset, forming pairs of image-based and text-based questions for common sense reasoning in the context of VQA. Indeed, text information cannot always be fully converted to images, and vice versa. It falls under the category of Translation Variant (TV) tasks.
Our data samples can be seen in Appendix \ref{app:VQA}

\subsection{Results}
\label{sec:res}

The main results of our evaluations across six different datasets are presented in Table \ref{tab:main}. Notably, for Translation Invariant (TI) tasks, where the input contains an equal amount of information for solving the task, we can observe significant differences between image and text input formats when tasks are challenging and require more reasoning processes. 

\textbf{Knowledge Access and Common Sense Reasoning} These tasks are relatively straightforward and don't require extensive logical reasoning. They appear to be simpler, with both images and text achieving accuracies of approximately 80-90\%. There is a high level of agreement between the vision and language modalities. In these cases, the consistency scores are notably high, with text input only slightly outperforming images in terms of accuracy. 

\textbf{Math Equation Solving, Math reasoning and State Machine Reasoning.} However, as tasks become more challenging and involve higher levels of logical reasoning, such as solving mathematical equations and engaging in reasoning processes, we observe a substantial difference of over 40\% in accuracy when comparing the performance of these tasks in image format and text formats. Consistency scores also exhibit a significant drop for these tasks (as low as 58\%). Notably, the model performs significantly better when presented with text-based questions, even though the essential information required to solve the task is equally available in the images. This highlights the considerable inconsistency in task-solving across modalities and underscores the superiority of the model's ability in one modality (Language) over the other (Vision).

\textbf{VQA Common Sense Reasoning.} Surprisingly, the model performs significantly better when prompted with text questions than with images in Translation Variant tasks, even though images often contain more information to answer the questions. This further highlights the large disparities between modalities (vision vs language) in certain complex tasks of these models. 
\begin{figure*}[t!]
    \centering
    \includegraphics[width=0.85\linewidth]{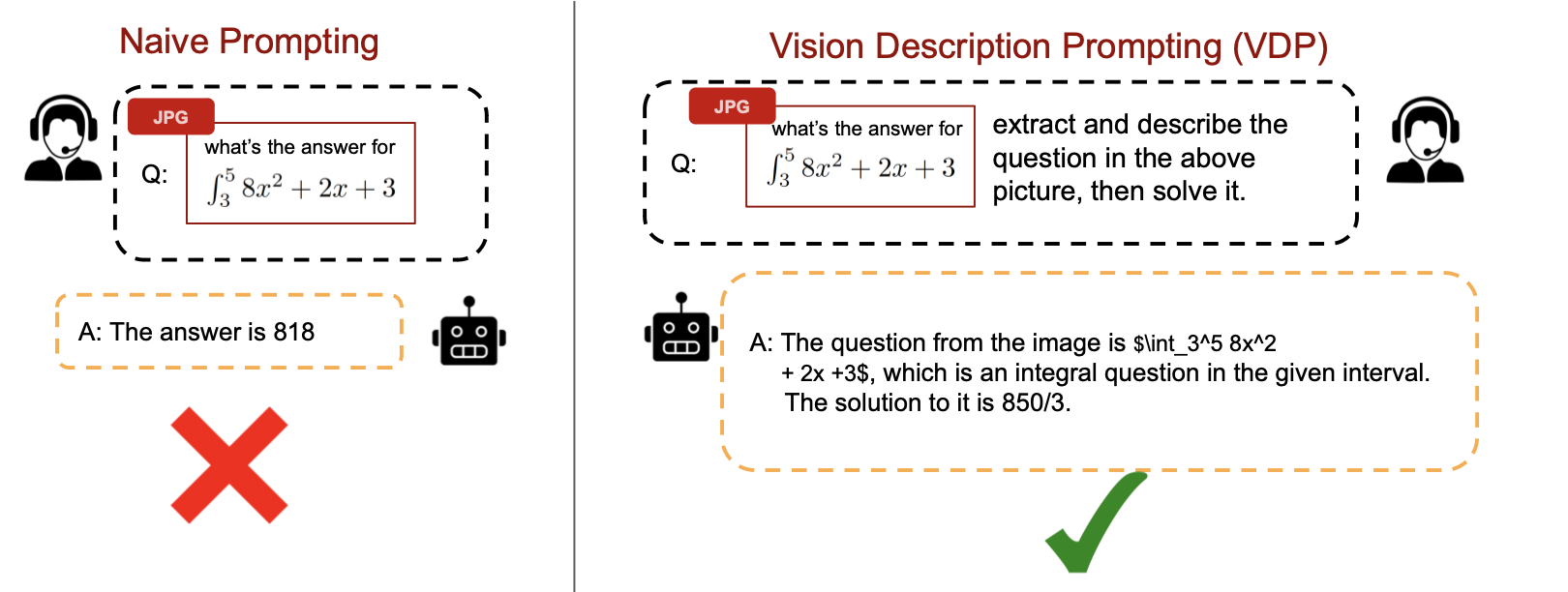}
    \caption{VDP method overview. The left part illustrates the naive way of prompting the vision tasks.  }
    \label{fig:VDP}
\end{figure*}

In conclusion, in multimodal systems like GPT-4V, the language modality demonstrates a \textbf{dominant} advantage when tasks are tackled in text format, despite the presence of the same or even more information in image format. This strongly suggests a non-compound modality network, where each modality exhibits varying levels of task-solving and reasoning capabilities.

Our hypothesis is that GPT-4V has not learned a highly consistent joint representation for vision and language, leading it to solve tasks separately depending on the input format. Given that GPT-4V leans more towards being a language model than a vision model, its strong biases are inclined towards language modeling, and this proficiency is not equally shared with vision-related tasks.

\subsection{Ablation study on text extraction from image.}
\label{sec:abl}
For several of the Translation Invariant (TI) tasks we employed, there may be a concern that differences in performance could arise from the model's inability to successfully extract or recognize information from images, even when the image contains the exact same content as the text (including formulas, questions, and choices).

To address this concern, we conducted an ablation study in which we simply extracted the text-format question from the images within our dataset. Specifically, we evaluated the model's ability to extract mathematical equation-solving questions, which are particularly challenging due to their complex symbols.

The results, with an impressive 87\% extraction accuracy achieved on mathematical equation-solving images, clearly illustrate GPT-4V's proficiency in extracting question information from the image part of our task dataset. This performance underscores its excellent capability in recognizing and extracting information from the image context.

Hence, it becomes evident that any performance disparity observed in these tasks can be primarily attributed to differences in the model's internal network reasoning between the vision modality and text, rather than being caused by other external factors such as incomplete information in images or problem of recognition information from the images. 

\section{Inspired Vision-depicting-prompting (VDP) }
As shown in Section \ref{sec:res}, for the same problem, LVLMs such as GPT4V performs much better when questions are presented in text format, even when information can be completely extracted and retained from the images instances. Therefore, we make use of such findings and propose the method of Vision-depicting-prompting (VDP) for improving reasoning ability through image context. 
\subsection{Prompting Details}
In the case of a task instance presented in image format, VDP diverges from directly soliciting an answer solely based on the image input, as illustrated in Figure \ref{fig:VDP}. Instead, we adopt a two-step process. We first prompt the model to extract and articulate the description of the image task using textual language. This approach aims to maximize the transformation of the image signal into a text signal, recognizing the inherently stronger reasoning abilities associated with text information, as demonstrated earlier. Subsequently, we prompt the model to provide an answer, taking into account both the text description of the task and the original image input, as depicted in Figure \ref{fig:VDP}. 

Unlike previous research that sought to enhance the reasoning abilities of multimodal models by augmenting input images with supplementary text~\cite{lin2022revive,hu2023promptcap}, VDP does not focus on information augmentation. This is particularly relevant for the TI task, where images already encapsulate all the necessary information required to complete the task. Hence, converting these images to text doesn't offer any additional insights. Instead, VDP is rooted in the observation that textual signals can significantly stimulate a model's reasoning capability, even when some information is missing from the visual context. VDP achieves this by explicitly extracting textual information from the images, thereby directly engaging the model's language processing capabilities more effectively. 

\begin{table}
\centering
\resizebox{\linewidth}{!}{%
\begin{tabular}{c|cccc}
\hline
\textbf{DataSet} & \textbf{Modality} & \textbf{Prompting} & \textbf{Accuracy} & \textbf{Consistency} \\
\hline\hline
 \multirow{3}{*}{MES}& text & naive & 0.80 & ---- \\ \cline{2-5}

 & \multirow{2}{*}{image} & naive  & 0.58 & 0.62\\

 & & VDP & \textbf{0.66} & \textbf{\ \ \ 0.86 $\Uparrow$} \\
\hline\hline

\multirow{3}{*}{MR}& text & naive &   0.96  & ---- \\ \cline{2-5}

 & \multirow{2}{*}{image} & naive  & 0.54 & 0.58\\

 & & VDP &  \textbf{\ \ \ \ 0.96 $\Uparrow$ } & \textbf{ \ \ 0.96 $\Uparrow$} \\
\hline
\end{tabular}
}

\caption{Result of VDP prompting. MES stands for Math Equation Solving and MR stands for Math Reasoning. $\Uparrow$ represents the improvement of more than 10\%. }
\label{tab:VDP}
\end{table}

\subsection{Experiment results for VDP}
We employed VDP on both the Math Equation Solving dataset and the Math Reasoning Dataset. These datasets previously exhibited subpar performance when using a simplistic prompting approach, as previously demonstrated. The outcomes are detailed in Table \ref{tab:VDP}. Remarkably, we observed a significant improvement in accuracy (by over 30\%) when solving problems based on image inputs. Furthermore, there was a substantial increase in the consistency score compared to prompting only using text. These results not only underline the effectiveness of our approach but also reinforce our hypothesis that models exhibit varied reasoning capabilities across different modalities. Properly utilizing such disparities can help to improve the performance in solving the tasks.

\section{Conclusion}
In this study, we performed a systematic analysis of the consistency across modalities in multimodal systems. Our results demonstrate that models like GPT-4V maintain a relatively independent internal representation of reasoning between visual and textual signals, as evidenced by our specially designed dataset. Notably, GPT-4V exhibits superior performance in language modeling compared to reasoning within a visual context. These findings offer valuable insights into the potential applications of such multimodal systems and highlight the need for more integrated system designs. Furthermore, we introduce a prompting solution to address this disparity.

\bibliography{custom}
\bibliographystyle{acl_natbib}

\clearpage
\onecolumn  
\section*{Appendix}
\appendix

\label{sec:appendix}

\section{Math Equation Solving Dataset Samples}
\label{app:MES}

\begin{figure*}[ht]
    \centering
    \includegraphics[width=0.7\linewidth]{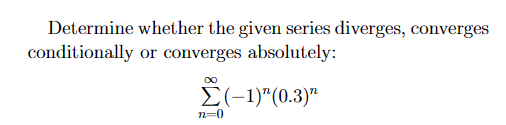}
    \caption{Sample 1 of Math Equation Dataset: Image.  }
    \label{fig:ME1}
\end{figure*}

\begin{table*}[ht]
    \centering
    
    \resizebox{\textwidth}{!}
    {%
    \begin{tabular}{|lp{1\linewidth}|}

    \hline
      \textbf{Text:}&
\texttt Determine whether the given series diverges, converges conditionally or converges absolutely:
      \\ &\$\$
        \textbackslash sum\_\{n=0\}\^ \ \{\textbackslash infty\}(-1)\^ \ n(0.3)\^ \ n\$\$
    \\ 
   \hline \end{tabular} 
    }

    \caption{Sample 1 of Math Equation Dataset: Text  
    }
    \label{table:ME1}
    
\end{table*}

\clearpage

\begin{figure*}[ht]
    \centering
    \includegraphics[width=0.52\linewidth]{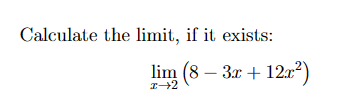}
    \caption{Sample 2 of Math Equation Dataset: Image.  }
    \label{fig:ME2}
\end{figure*}

\begin{table*}[h]
    \centering
    
    \resizebox{0.95\textwidth}{!}
    {%
    \begin{tabular}{|lp{1\linewidth}|}

    \hline
      \textbf{Text:}&
\texttt  Calculate the limit, if it exists:
\$\$ \textbackslash lim\_\{ x \textbackslash rightarrow 2 \} \textbackslash left (8-3 x+12 x\^ \ 2 \textbackslash right)\$\$
    \\ 
   \hline \end{tabular} 
    }

    \caption{Sample 2 of Math Equation Dataset: Text  
    }
    \label{table:ME2}
    
\end{table*}

\newpage

\section{Math Reasoning Dataset}
\label{app:MR}
\begin{figure*}[ht]
    \centering
    \includegraphics[width=0.8\linewidth]{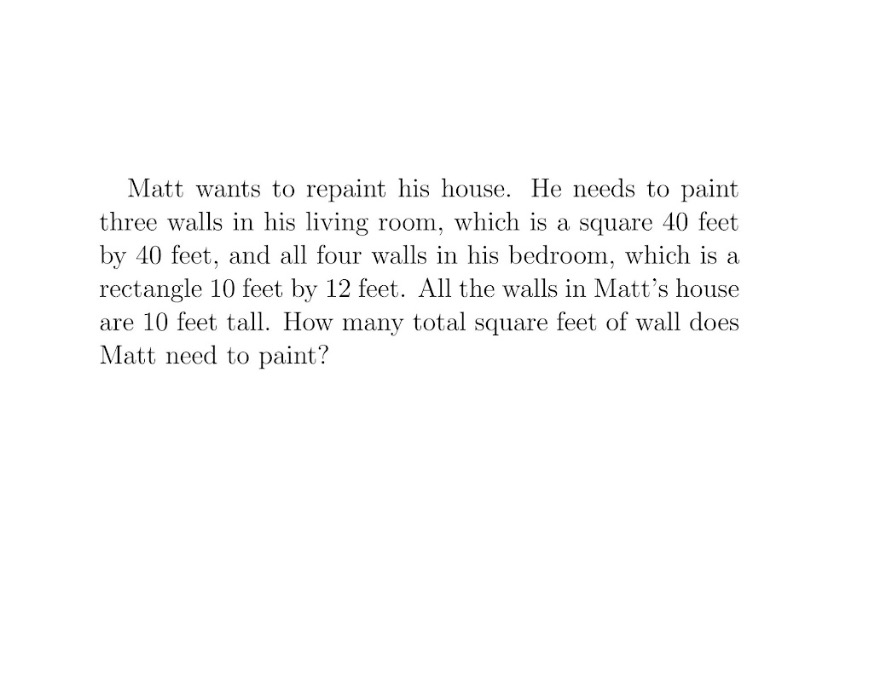}
    \caption{Sample 1 of Math Reasoning Dataset: Image.  }
    \label{fig:MR1}
\end{figure*}

\begin{table*}[ht]
    \centering
    
    \resizebox{0.95\textwidth}{!}
    {%
    \begin{tabular}{|lp{1\linewidth}|}

    \hline
      \textbf{Text:}&Matt wants to repaint his house. He needs to paint three walls in his living room, which is a square 40 feet by 40 feet, and all four walls in his bedroom, which is a rectangle 10 feet by 12 feet. All the walls in Matt's house are 10 feet tall. How many total square feet of wall does Matt need to paint?
    \\ 
   \hline \end{tabular} 
    }

    \caption{Sample 1 of Math Reasoning Dataset: Text  
    }
    \label{table:MR1}
    
\end{table*}

\clearpage
\begin{figure*}[ht]
    \centering
    \includegraphics[width=0.7\linewidth]{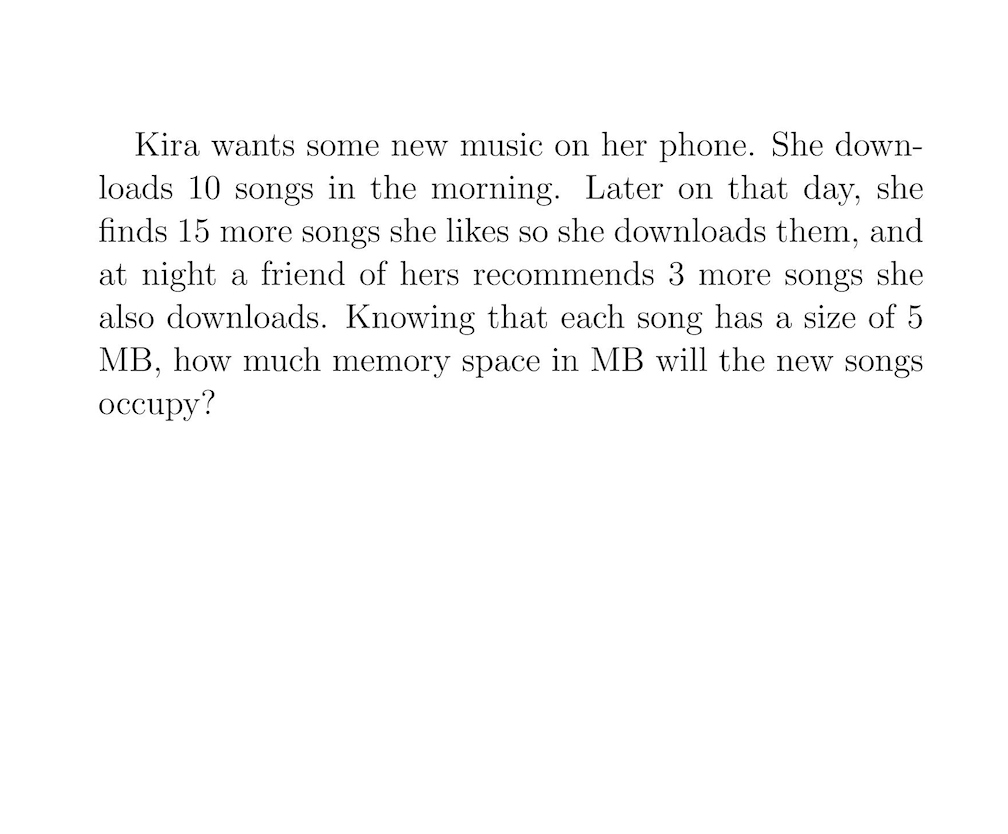}
    \caption{Sample 2 of Math Reasoning Dataset: Image.  }
    \label{fig:MR2}
\end{figure*}

\begin{table*}[ht]
    \centering
    
    \resizebox{0.95\textwidth}{!}
    {%
    \begin{tabular}{|lp{1\linewidth}|}

    \hline
      \textbf{Text:}&Kira wants some new music on her phone. She downloads 10 songs in the morning. Later on that day, she finds 15 more songs she likes so she downloads them, and at night a friend of hers recommends 3 more songs she also downloads. Knowing that each song has a size of 5 MB, how much memory space in MB will the new songs occupy?
    \\ 
   \hline \end{tabular} 
    }

    \caption{Sample 2 of Math Reasoning Dataset: Text  
    }
    \label{table:MR2}
    
\end{table*}
\clearpage
\section{Knowledge Access Dataset}
\label{app:knowledge}

\begin{figure*}[ht]
    \centering
    \includegraphics[width=0.7\linewidth]{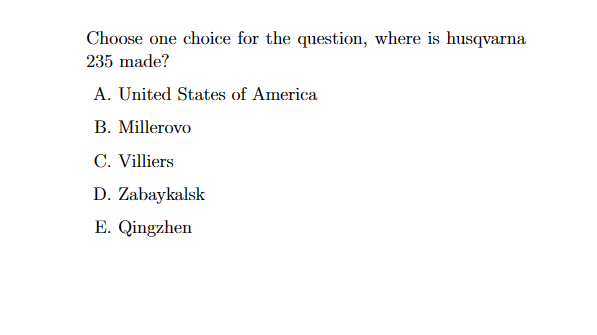}
    \caption{Sample 1 of Knowledge Access Dataset: Image.  }
    \label{fig:KA1}
\end{figure*}

\begin{table*}[ht]
    \centering
    
    \resizebox{0.95\textwidth}{!}
    {%
    \begin{tabular}{|lp{1\linewidth}|}

    \hline
      \textbf{Text:}&Choose one choice for the question, where is husqvarna 235 made?
      
     A. United States of America
     B. Millerovo
     C. Villiers
     D. Zabaykalsk
     E. Qingzhen
    \\ 
   \hline \end{tabular} 
    }

    \caption{Sample 1 of Knowledge Access Dataset: Text  
    }
    \label{table:KA1}
    
\end{table*}

\clearpage
\begin{figure*}[ht]
    \centering
    \includegraphics[width=0.7\linewidth]{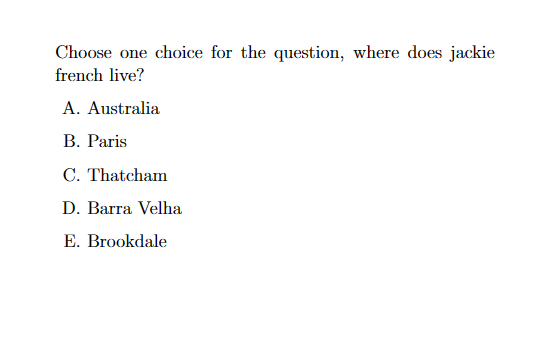}
    \caption{Sample 2 of Knowledge Access Dataset: Image.  }
    \label{fig:KA2}
\end{figure*}

\begin{table*}[ht]
    \centering
    
    \resizebox{0.95\textwidth}{!}
    {%
    \begin{tabular}{|lp{1\linewidth}|}

    \hline
      \textbf{Text:}&Choose one choice for the question, where does jackie french live?

 A. Australia
 B. Paris
 C. Thatcham
 D. Barra Velha
 E. Brookdale
    \\ 
   \hline \end{tabular} 
    }

    \caption{Sample 2 of Knowledge Access Dataset: Text  
    }
    \label{table:KA2}
    
\end{table*}

\clearpage
\section{State Machine Dataset}
\label{app:state}

\begin{figure*}[ht]
    \centering
    \includegraphics[width=0.6\linewidth]{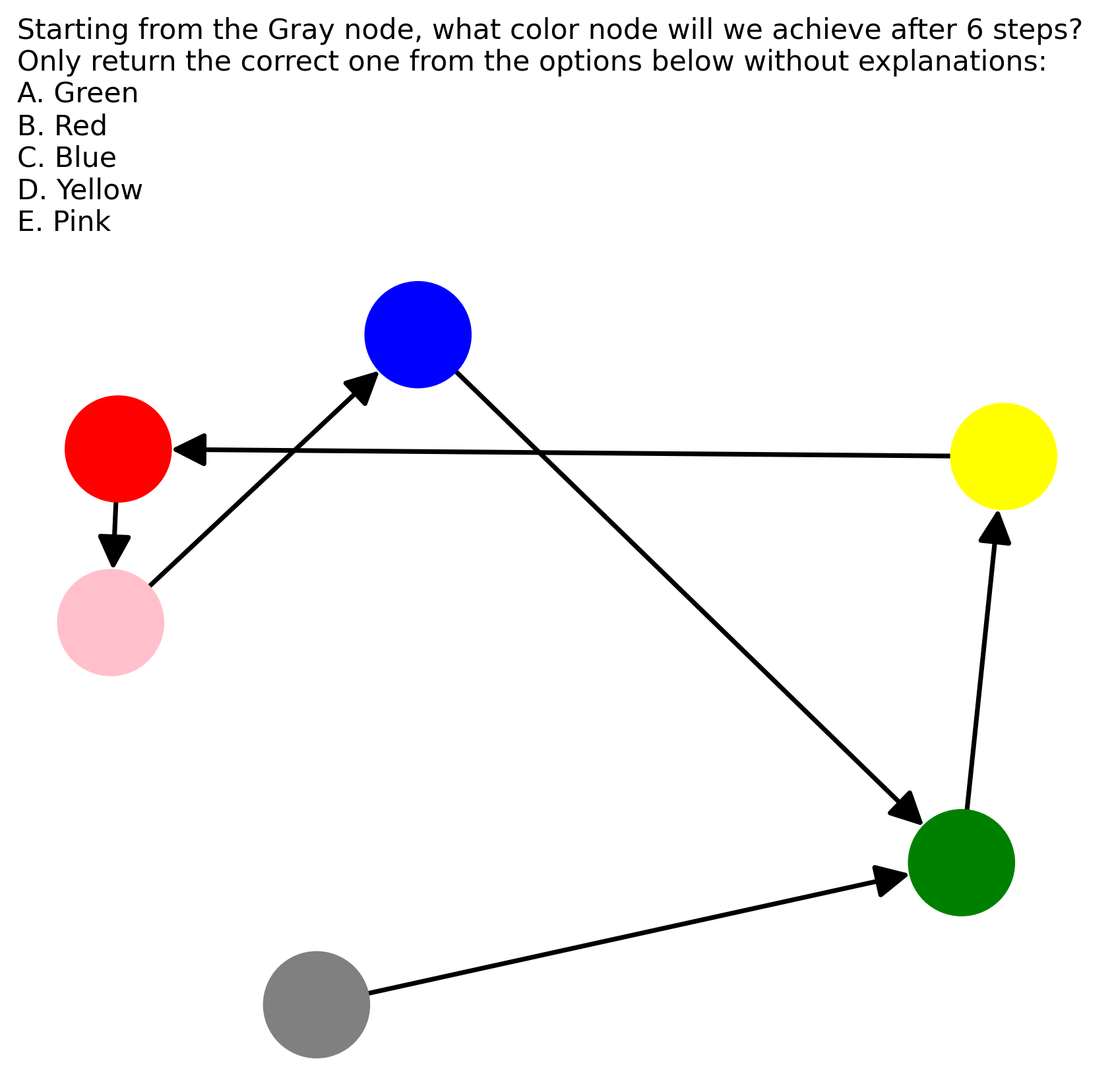}
    \caption{Sample 1 of State Machine Dataset: Image.  }
    \label{fig:SM1}
\end{figure*}

\begin{table*}[ht]
    \centering
    
    \resizebox{0.95\textwidth}{!}
    {%
    \begin{tabular}{|lp{1\linewidth}|}

    \hline
      \textbf{Text:}&Consider a graph with the following directed edges: 
Yellow leads to Red;
Green leads to Yellow;
Red leads to Pink;
Blue leads to Green;
Gray leads to Green;
Pink leads to Blue.
Starting from the Gray node, what color node will we achieve after 6 steps?
Only return the correct one from the options below without explanations:
A. Green
B. Red
C. Blue
D. Yellow
E. Pink
    \\ 
   \hline \end{tabular} 
    }

    \caption{Sample 1 of State Machine Dataset: Text  
    }
    \label{table:SM1}
    
\end{table*}

\clearpage
\begin{figure*}[ht]
    \centering
    \includegraphics[width=0.6\linewidth]{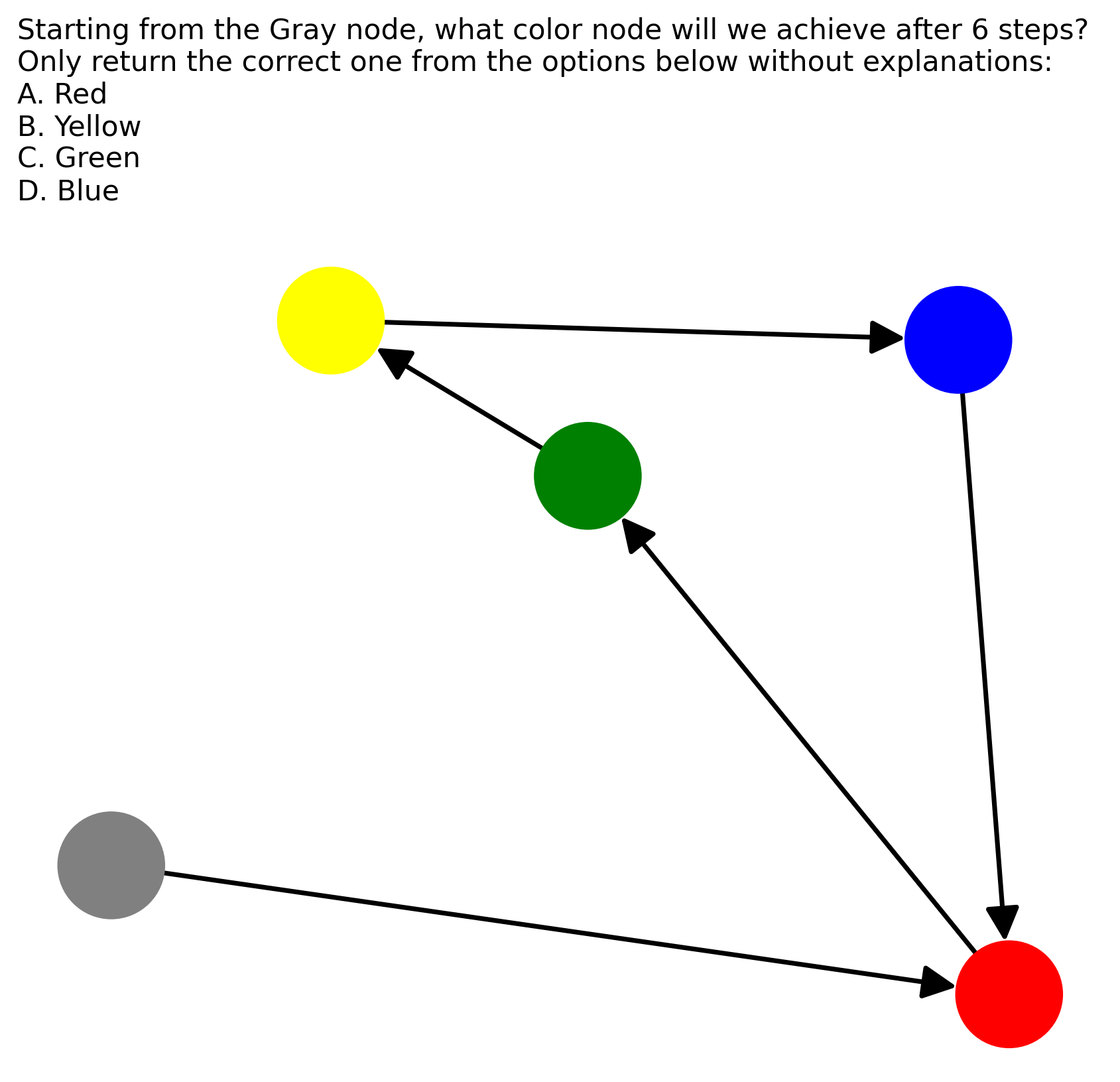}
    \caption{Sample 2 of State Machine Dataset: Image.  }
    \label{fig:SM2}
\end{figure*}

\begin{table*}[ht]
    \centering
    
    \resizebox{0.95\textwidth}{!}
    {%
    \begin{tabular}{|lp{1\linewidth}|}

    \hline
      \textbf{Text:}&Consider a graph with the following directed edges: 
Gray leads to Red;
Yellow leads to Blue;
Blue leads to Red;
Red leads to Green;
Green leads to Yellow.
Starting from the Gray node, what color node will we achieve after 6 steps?
Only return the correct one from the options below without explanations:
A. Red
B. Yellow
C. Green
D. Blue
    \\ 
   \hline \end{tabular} 
    }

    \caption{Sample 2 of State Machine Dataset: Text  
    }
    \label{table:SM2}
    
\end{table*}

\clearpage
\section{Common Sense Reasoning Dataset}

\label{app:common}

\begin{figure*}[ht]
    \centering
    \includegraphics[width=0.7\linewidth]{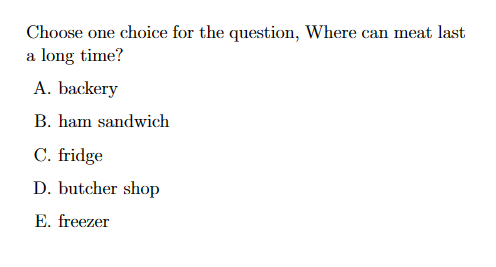}
    \caption{Sample 1 of Common Sense Reasoning Dataset: Image.  }
    \label{fig:CSR1}
\end{figure*}

\begin{table*}[ht]
    \centering
    
    \resizebox{0.95\textwidth}{!}
    {%
    \begin{tabular}{|lp{1\linewidth}|}

    \hline
      \textbf{Text:}&Choose one choice for the question, Choose one choice for the question, Where can meat last a long time?
 A. backery
 B. ham sandwich
 C. fridge
 D. butcher shop
 E. freezer
    \\ 
   \hline \end{tabular} 
    }

    \caption{Sample 1 of Common Sense Reasoning Dataset: Text  
    }
    \label{table:CSR1}
    
\end{table*}

\clearpage
\begin{figure*}[ht]
    \centering
    \includegraphics[width=0.7\linewidth]{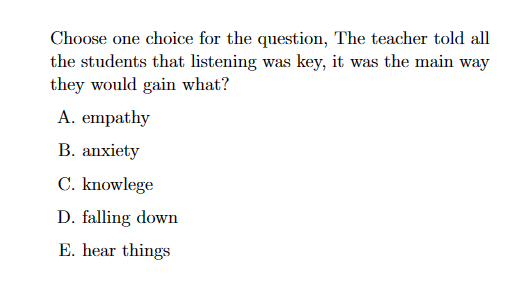}
    \caption{Sample 2 of Common Sense Reasoning Dataset: Image.  }
    \label{fig:CSR2}
\end{figure*}

\begin{table*}[ht]
    \centering
    
    \resizebox{0.95\textwidth}{!}
    {%
    \begin{tabular}{|lp{1\linewidth}|}

    \hline
      \textbf{Text:}&Choose one choice for the question, The teacher told all the students that listening was key, it was the main way they would gain what?

 A. empathy
 B. anxiety
 C. knowlege
 D. falling down
 E. hear things
    \\ 
   \hline \end{tabular} 
    }

    \caption{Sample 2 of Common Sense Reasoning Dataset: Text  
    }
    \label{table:CSR2}
    
\end{table*}

\clearpage
\section{VQA Common Sense Reasoning Dataset}
\label{app:VQA}

\begin{figure*}[ht]
    \centering
    \includegraphics[width=0.8\linewidth]{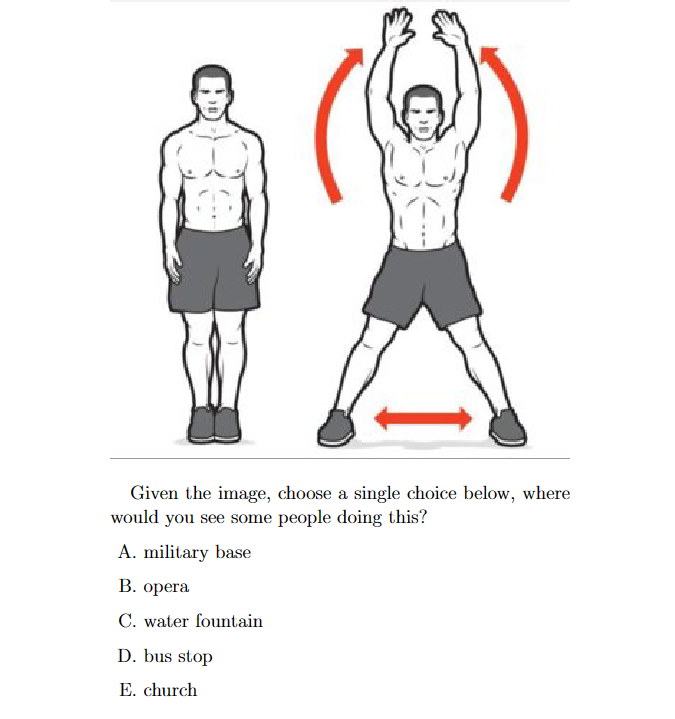}
    \caption{Sample 1 of VQA Dataset: Image.  }
    \label{fig:VQA1}
\end{figure*}

\begin{table*}[ht]
    \centering
    
    \resizebox{0.95\textwidth}{!}
    {%
    \begin{tabular}{|lp{1\linewidth}|}

    \hline
      \textbf{Text:}&Choose one choice for the question, Where would you see some people doing jumping jacks?

 A. military base
 B. opera
 C. water fountain
 D. bus stop
 E. church
    \\ 
   \hline \end{tabular} 
    }

    \caption{Sample 1 of VQA Dataset: Text  
    }
    \label{table:VQA1}
    
\end{table*}

\newpage

\begin{figure*}[ht]
    \centering
    \includegraphics[width=0.8\linewidth]{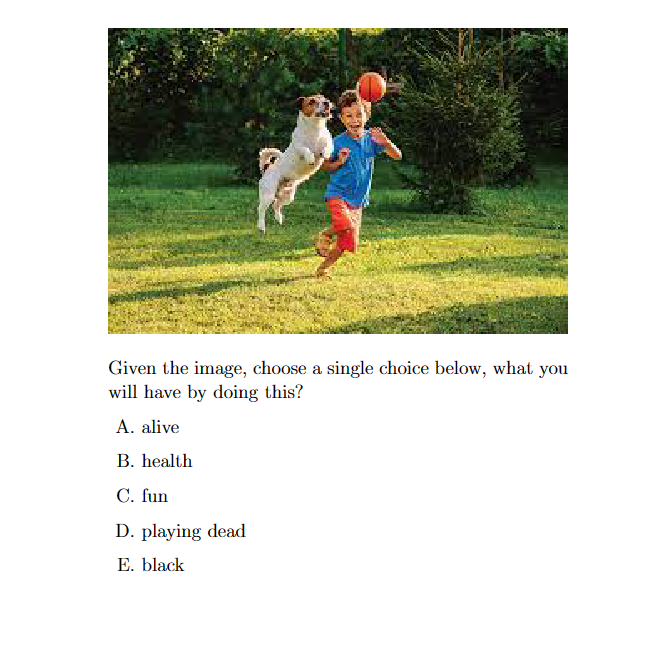}
    \caption{Sample 2 of VQA Dataset: Image.  }
    \label{fig:VQA2}
\end{figure*}

\begin{table*}[ht]
    \centering
    
    \resizebox{0.95\textwidth}{!}
    {%
    \begin{tabular}{|lp{1\linewidth}|}

    \hline
      \textbf{Text:}&Choose one choice for the question, When you play around with your dog they will have?

 A. alive
 B. health
 C. fun
 D. playing dead
 E. black
    \\ 
   \hline \end{tabular} 
    }

    \caption{Sample 2 of VQA Dataset: Text  
    }
    \label{table:VQA2}
    
\end{table*}

\end{document}